\documentclass[runningheads]{llncs}
\usepackage[T1]{fontenc}
\usepackage{graphicx}
\usepackage{booktabs}
\usepackage[misc]{ifsym}
\newcommand{\corr}{(\Letter)}
\usepackage{mwe}
\usepackage{booktabs}  
\usepackage{tabularx}  

\usepackage{hyperref}

\begin{document}

\title{Towards Explainable Deep Clustering\\ for Time Series Data}

\titlerunning{Towards Explainable Deep Clustering for Time Series Data}

\author{Udo Schlegel\inst{1,2}\orcidID{0000-0002-8266-0162}\corr \and
Gabriel Marques Tavares\inst{1,2}\orcidID{0000-0002-2601-8108} \and
Thomas Seidl\inst{1,2}\orcidID{0000-0002-4861-1412}}

\authorrunning{Schlegel et al.}

\institute{LMU Munich, Germany
\and
Munich Center for Machine Learning (MCML), Germany\\
\email{udo.schlegel@lmu.de}}

\maketitle              

\begin{abstract}
Deep clustering enables uncovering hidden patterns and groups in complex time series data.
Yet, its opaque decision-making limits use in safety-critical settings. 
This review survey offers a structured overview of explainable deep clustering for time series, collecting current methods and their real-world applications.
We discuss and compare peer-reviewed and preprint papers through application domains across healthcare, finance, IoT, and climate science.
Our analysis reveals that most work relies on autoencoder and attention architectures, with limited support for streaming, irregularly sampled, or privacy-preserved series, and interpretability is still primarily treated as an add-on and not a goal.
To push the field forward, we outline six research opportunities: 
(1) combining complex networks with built-in interpretability; 
(2) setting up clear, faithfulness-focused evaluation metrics for unsupervised explanations; 
(3) building explainers that scale to large data and adapt to live data streams; 
(4) crafting explanations tailored to specific domains; 
(5) adding human-in-the-loop methods that refine clusters and explanations together; and
(6) improving our understanding of how time series clustering models work internally. 
By making interpretability a primary design goal rather than an afterthought, we propose the groundwork for the next generation of trustworthy deep clustering time series analytics.


\keywords{Explainable AI \and Deep Clustering \and Time Series Data}
\end{abstract}

\section{Introduction \& Motivation}

Time series clustering is an unsupervised learning task that groups similar temporal sequences without prior labels. 
Traditional approaches often rely on distance measures (e.g., dynamic time warping or Euclidean distance) and yield clusters based on raw time series similarity. 
However, these methods struggle with high-dimensional, noisy, or complex time series data. 
In recent years, deep learning-based clustering has emerged to tackle these challenges. 
Deep clustering methods use neural networks (such as autoencoders, RNNs, or transformers) to learn compressed representations (embeddings) of time series, and then perform clustering in the latent space or even include clustering in the learning process. 
This can capture non-linear temporal patterns and improve clustering quality compared to classical techniques~\cite{paparrizos_bridging_2024}.
For example, methods such as Deep Embedded Clustering (DEC)~\cite{xie_unsupervised_2016} and its variants have been extended to time series by combining reconstruction losses with clustering objectives~\cite{alqahtani_deep_2021}.
These deep clustering models have achieved notable success in grouping time series data (e.g., sensor readings, human activity signals) into meaningful clusters~\cite{alqahtani_deep_2021}.

Despite improved performance, a key limitation is the black-box nature of deep models. 
The latent features learned by deep networks are often not easily interpretable, making it hard to understand why certain time series samples are clustered together~\cite{bonifati_time2feat_2022}.
In critical domains (healthcare, finance), human experts need to comprehend the clustering results and trust the model’s behavior to apply these to their use cases. 
As Bonifati et al.~\cite{bonifati_time2feat_2022} note, many systems “maximize effectiveness… but fail to guarantee the interpretability of the results,” which hinders their use in real-world scenarios where human oversight is required.
This has led to growing interest in Explainable Deep Clustering for Time Series, i.e., methods that integrate explainability into the deep clustering pipeline. 

This paper provides a structured overview, review, and research opportunities for explainable deep clustering of time series data.
We collected and systematised peer-reviewed and high-quality preprint studies over the last few years.
Our comparison table and cross-domain case studies, covering healthcare, finance, IoT, and climate science, reveal that today’s landscape remains dominated by autoencoder-plus-attention designs, rarely addresses streaming, privacy-preserved, or highly irregular data, and targets explainability as an add-on.
Building on these observations, we articulate six research opportunities to highlight important directions: (1) coupling complex models with intrinsic interpretability; (2) defining faithfulness-centred evaluation metrics for unsupervised explanations; (3) creating explainers that evolve with live data; (4) tailoring explanations to domain semantics; (5) integrating human-in-the-loop refinement; and (6) advancing mechanistic interpretability of learned temporal features.

\section{Explainability in Deep Learning}

Explainable AI refers to techniques that make the decisions of complex models understandable to humans. 
In supervised learning (classification/regression), popular XAI methods include feature attribution (e.g., LIME~\cite{ribeiro_why_2016}, SHAP~\cite{lundberg_unified_2017}), which assigns importance scores to input features, attention mechanisms that highlight influential parts of the input, prototype- or example-based explanations that show representative cases, and visualizations of internal model workings~\cite{theissler_explainable_2022}. 
However, time series pose particular challenges for explainability~\cite{schlegel_towards_2019}. 
Time series inputs (especially multivariate sequences) are often high-dimensional and not immediately interpretable (unlike images or text)~\cite{theissler_explainable_2022}.
Only limited XAI research has focused on time series data compared to vision or NLP~\cite{vielhaben_explainable_2024}.
Typical explanation strategies for time series include identifying important time segments or sensor channels, transforming data into interpretable domains (e.g., frequency spectra), or finding prototypical patterns~\cite{theissler_explainable_2022}.

For unsupervised learning, such as clustering, explainability is even more challenging. 
There are no pre-defined class labels or ground-truth explanations for clusters, and clusters may reflect subtle patterns in the data~\cite{ren_deep_2024,zhou_comprehensive_2024}. 
The goal of explainable clustering is to describe or justify why a cluster exists – e.g., by summarizing common patterns in a cluster, or highlighting what differentiates one cluster from others~\cite{hu_interpretable_2024}.
Previous works in explainable cluster analysis have proposed taxonomies of techniques: (1) Pre-clustering explainability (data transformations to make clustering more interpretable), (2) In-clustering explainability (algorithms that build interpretability into the clustering process), and (3) Post-hoc explainability (analyzing the results after clustering to extract explanations)~\cite{hu_interpretable_2024}.
In the context of deep clustering, all three approaches are relevant. 
For further information on deep clustering approaches and methods, we refer to Ren et al.~\cite{ren_deep_2024} and Zhou et al.~\cite{zhou_comprehensive_2024}.
Next, we examine methods that specifically combine deep learning for time series clustering with explainability techniques or mention these during the approach description.

\section{Combining XAI with Deep Clustering for Time Series}

Achieving explainability in deep time series clustering has been approached from multiple angles. 
Below, we survey representative methods, roughly grouped by the explainability strategy they employ or mention in their description.

\subsection{Attention-Based Approaches}

One way to introduce interpretability is by using attention mechanisms in the clustering model. 
Attention layers can weigh the importance of different time steps or features, effectively pointing to which parts of the time series are most influential in determining the cluster assignment. 
Ienco and Interdonato~\cite{ienco_deep_2020} propose DeTSEC (Deep Time Series Embedding Clustering) as an example of this approach. 
DeTSEC first utilizes a gated recurrent autoencoder with an attention mechanism to generate a preliminary embedding for each multivariate time series. 
The attention weights highlight subsequences that the network deems important. 
In a second stage, DeTSEC refines the embeddings with a clustering-oriented loss to improve cluster separation. 
By inspecting the learned attention weights, one can identify which time points or sensors contributed most to a series’ embedding and cluster, providing a form of explanation for the clustering. 
For instance, if a certain peak or pattern in the time series consistently gets high attention for all members of a cluster, that pattern can be presented as a distinguishing feature of that cluster. 
Attention-based deep clustering has demonstrated success on varied time series data (e.g., speech signals and human activity data), outperforming other methods while offering some interpretability of the learned representation. 
Attention is not a perfect explanation (it indicates correlation rather than causation), but it serves as an intrinsic interpretability tool in many deep models using an attention mechanism~\cite{bibal_is_2022}.

Beyond RNN-based attention, emerging work uses transformer architectures for time series classification~\cite{theissler_explainable_2022}, which inherently utilize multi-head self-attention and can also be applied to clustering~\cite{nguyen_clusformer_2021}. 
While primarily aimed at improving accuracy, these models can also be probed for explainability: the attention scores over time steps or variables can be visualized to see what the model “attended” to when forming clusters~\cite{bibal_is_2022}. 
For example, a transformer-based clustering model might attend to seasonal spikes or anomalies in a time series that drive clustering for a specific cluster. 
However, research explicitly combining transformer time series clustering with human-friendly explanations is still nascent (most current transformer models focus on classification or forecasting)~\cite{theissler_explainable_2022}.

Based on the taxonomy proposed by Hu et al.~\cite{hu_interpretable_2024}, we argue that attention-based approaches predominantly fall within the category of in-clustering explanations. 
This is because, at each step of the clustering process, attention weights inherently provide a means of interpretation: they quantify the relative importance of input features or tokens, thereby yielding explanations that are both immediate and automatically generated as part of the model’s operation, rather than requiring a separate post-hoc procedure.

\subsection{Prototype and Example-Based Explanations}

Another line of research provides explanations by prototypes or examples for learned pattern representations at each cluster. 
The idea is to characterize clusters in terms of human-understandable exemplars or features. 
One prominent approach is the use of shapelets, originally developed for classification, which are short representative subsequences that are maximally characteristic of a class~\cite{ye_time_2009}.
Shapelets have long been used in time series classification as interpretable features~\cite{theissler_explainable_2022}, and more recently have been applied to clustering.
For example, El Amouri et al.~\cite{elamouri_constrained_2023} introduce Constrained DTW-Preserving Shapelets (CDPS) for explainable time series clustering.
In CDPS, the model learns a set of shapelet patterns that capture the variability in the dataset while approximately preserving Dynamic Time Warping distances between series~\cite{elamouri_constrained_2023}.
The learned shapelets form a new feature space for clustering, which is inherently interpretable: 
each dimension corresponds to the DTW similarity of a time series to a certain shapelet (a meaningful pattern), forming a new similarity vector~\cite{elamouri_constrained_2023}.
After clustering in this space, they propose Shapelet Cluster Explanation (SCE) methods that output a small set of shapelets being most descriptive for time series samples in the cluster, summarizing inherent patterns~\cite{elamouri_constrained_2023}.
In essence, a cluster can be explained by saying “time series in this group all contain pattern A and lack pattern B,” where A and B are actual subsequences (such as a particular waveform shape) that can be visualized.
This approach yields explanations that are domain-meaningful (experts can interpret the shapelet patterns using their domain knowledge). 
A limitation is that shapelet-based methods require extensive search or complex training, but they provide a clear link between cluster assignment and understandable raw data patterns.

Prototype-based explanations need not be limited to subsequences. 
Some deep clustering methods present whole-series prototypes or representative examples for each cluster. 
A simple form is to identify medoids or centroids in the original space – e.g., select one actual time series per cluster that is closest to the cluster center. This selected series can act as a prototype that “stands for” the cluster. 
More sophisticated is learning prototypes through the model itself. 
For instance, autoencoder clustering models can be used to decode cluster centroids from the latent space back into the time domain. 
A variational autoencoder (VAE) with a clustering prior (like a Gaussian mixture in latent space~\cite{dilokthanakul_deep_2016}) can generate a prototype time series from each cluster’s latent mean. 
These prototype series can then be examined by humans (or matched to real examples) for interpretation. 
Such an approach was implicitly used by VAE+SOM-based models, where each neuron on the self-organizing map corresponded to a prototype time series pattern in the input space~\cite{fortuin_som_2019}. 
Generally, prototype and exemplar strategies enhance explainability by grounding clusters in familiar examples or patterns rather than abstract coordinates.

Based on the taxonomy by Hu et al.~\cite{hu_interpretable_2024}, we argue that prototype- and example-based approaches heavily depend on the applied method to which category these correspond.
CDPS~\cite{elamouri_constrained_2023} fits the pre-clustering explainability as the approach already transforms the space into a new feature space using the shapelets.
However, VAE+SOM-based models such as Fortuin et al.~\cite{fortuin_som_2019} focus on changing the internal clustering approach by binding every prototype pattern to a neuron, creating a combination of pre-clustering and in-clustering explainability.

\subsection{Self-Organizing Maps and Topological Interpretability}

Integrating interpretable structures into deep clustering architectures is another powerful approach. 
Self-Organizing Maps (SOMs), a classic neuro-fuzzy method, project high-dimensional data onto a 2D grid of neurons while preserving topological relationships. 
Several recent works combine SOMs with deep learning to obtain the benefits of both: high clustering performance and an interpretable organization of clusters. 
Fortuin et al.~\cite{fortuin_som_2019} introduced SOM-VAE, which couples a VAE with a SOM in the latent space to learn discrete representations of time series.
The SOM imposes a two-dimensional grid structure on the latent embeddings, so each time series is mapped to a particular neuron (cluster) on a grid. 
This strongly enhances interpretability, as nearby clusters on the map correspond to similar time series patterns~\cite{paparrizos_bridging_2024}.
The grid can be visualized, and one can observe smooth transitions across the map, understanding how clusters relate to each other. 
Importantly, Fortuin et al.~\cite{fortuin_som_2019} also integrated a Markov transition model on the SOM to capture temporal dynamics (for sequences of series states), providing additional prediction capabilities and explainability in terms of state transition probabilities.

Building on SOM-VAE, Manduchi et al.~\cite{manduchi_dpsom_2019} developed DPSOM and its time series extension T-DPSOM~\cite{manduchi_t_2021}.
DPSOM uses a VAE with a probabilistic SOM loss to improve clustering quality without losing the SOM’s visualization capability~\cite{manduchi_dpsom_2019}. 
T-DPSOM further adds an LSTM-based temporal component, allowing not only clustering of multivariate time series but also forecasting in the latent space~\cite{manduchi_t_2021}. 
A key outcome is that T-DPSOM can produce interpretable visualizations of patient state trajectories on the SOM grid, along with uncertainty estimates~\cite{manduchi_t_2021}. 
In a healthcare setting (ICU patient monitoring), this means one can cluster patient time series data (e.g., vitals, lab measurements) into prototypical “states” and then track a patient’s progression through these states on a 2D map – a very intuitive explanation for clinicians~\cite{manduchi_t_2021}. 
The SOM-based deep clustering family thus offers an ante-hoc interpretability: the model is inherently structured to be interpretable by design (via the map), rather than explaining after the fact. 
These methods have demonstrated competitive clustering accuracy on benchmarks (e.g., SOM-VAE outperforms or matches other deep clustering methods on image and medical time series data~\cite{fortuin_som_2019}) while providing a user-friendly view of the clusters.

As already discussed in the previous section, the SOM-based explanations built in an in-clustering explainability and help to generate explanations while clustering.
All methods enhance the clustering approach while T-DPSOM~\cite{manduchi_t_2021} builds on SOM-VAE~\cite{fortuin_som_2019}.

\subsection{Post-hoc Interpretation and Visualization}

In addition to building interpretability into models, researchers have explored post-hoc explainability techniques applied to deep clustering results. 
One common strategy is to treat the cluster labels (obtained from the deep model) as pseudo-targets and train an auxiliary interpretable model to predict them. 
For example, one can fit a decision tree classifier that takes a time series (or engineered features of it) as input and predicts its cluster. The decision tree’s splits then reveal which features or thresholds distinguish the clusters. 
This idea is used in the CLAMP framework (Cluster Analysis with Multidimensional Prototypes) and related work, which derives rule-based explanations for clusters by supervised learning on the cluster assignments~\cite{bobek_enhancing_2022}. 
For time series, these features might be domain-specific (e.g., average heart rate, volatility of a financial time series, seasonal peak value). 
The decision rules provide an intelligible summary (e.g., “Cluster 1 consists of series where mean value > X and frequency of spikes < Y”). 
A caution is that such post-hoc models are approximations of the deep model’s behavior and may not capture all nuances in a trustworthy setting. 
Still, they offer an entry- and high-level understanding.

Model-agnostic explainers like LIME~\cite{ribeiro_why_2016} and SHAP~\cite{lundberg_unified_2017} can also be adapted for clustering. 
One approach is to use a similarity metric: given a target series and its cluster label, generate perturbations of that series, see how the clustering assignment changes (perhaps by feeding through the deep model), and then fit a local linear model. 
This would yield local feature importance for that single series’s clustering outcome. For instance, one could apply SHAP to an autoencoder+cluster network by inputting time series features (such as summary statistics or time-sampled points) to get Shapley values indicating which features “pushed” the series toward its cluster. 
While this has been explored less in the literature, it is a promising direction for instance-level explanations of cluster membership. 
In time series, one might interpret a positive SHAP value for a particular time segment or sensor reading as evidence that that aspect of the series is characteristic of the cluster.

Visualization remains a powerful tool for explaining clustering results. 
With deep models, one can visualize the learned embedding space to see the cluster structure. 
For example, applying t-SNE~\cite{maaten_visualizing_2008} or UMAP~\cite{mcinnes_umap_2018} on the latent representations from a deep model can produce a 2D plot where clusters form distinct groups~\cite{paparrizos_bridging_2024}.
Such plots help verify that the model has found a sensible separation and can be presented to end-users for insights (with points potentially colored by known attributes to interpret clusters). 
Additionally, specific visualization techniques have been devised for time series. 
One notable method is the Virtual Inspection Layer proposed by Vielhaben et al.~\cite {vielhaben_explainable_2024}.
Although aimed at classification, the concept is relevant: they insert a layer that transforms time series into an interpretable domain (like a Fourier spectrum) and then propagate explanation signals (using LRP) to that domain~\cite{vielhaben_explainable_2024}.
For clustering, one could imagine a similar approach where the time series are transformed (e.g., to the frequency domain or into a set of summary statistics) within the model, so that any relevance or distance can be explained in terms of those transformed features. 
Lastly, interactive visualization systems can allow users to explore cluster-wise average series, variance, and identified important segments (for example, highlighting sub-sequences with high attention or high contribution to cluster decisions, similar to attributions in supervised settings~\cite{schlegel_time_2021}). 
Such tools are increasingly important in domains where analysts need to validate clusters.

Here, as the title already suggests, the main focus of the explanation relies on the post-hoc explainability, either using post-hoc explanation methods or introducing possibilities to enable post-hoc explanations.

\begin{table}[t]
\caption{
Representative deep clustering methods for time series with explainability, grouped by the interpretable clustering three-stage taxonomy\cite{hu_interpretable_2024}. Method, model type, explanation approach, taxonomy stage, and example domains are shown. (Note: CLAMP is model-agnostic, shown here as an example of post-hoc cluster explanation.).}
\label{tab:deep_ts_xai}
\footnotesize                             
\setlength{\tabcolsep}{5pt}               
\renewcommand{\arraystretch}{1.15}        

\begin{tabularx}{\linewidth}{@{}%
      p{2.3cm}    
      p{2.0cm}    
      p{2.8cm}    
      p{1.4cm}    
      p{2.5cm}@{}}
\toprule
\textbf{Method (Year)} & \textbf{Model} & \textbf{Explainability} & \textbf{Stage} & \textbf{Presented Domain}\\
\midrule
SOM--VAE (2019)~\cite{fortuin_som_2019}            & VAE + 2-D SOM                     & Latent grid; prototype neurons                & In-cluster & ICU vitals; MNIST seq.\\
T-DPSOM (2020)~\cite{manduchi_t_2021}             & VAE + LSTM + prob.\ SOM           & Trajectory map; uncertainty                   & In-cluster & ICU patient states\\
DeTSEC (2020)~\cite{ienco_deep_2020}              & Attentive GRU autoencoder         & Attention on key subsequences                 & In-cluster & Speech, gesture, ECG\\
Time2Feat (2022)~\cite{bonifati_time2feat_2022}           & Feature extractor + DNN           & Interpretable domain features                 & Pre-cluster & 18 IoT / activity sets\\
CDPS Shapelets (2023)~\cite{elamouri_constrained_2023}      & CNN shapelet learner + $k$-means  & Representative subsequences (shapelets)       & Pre-cluster & UCR archive\\
Explain.\ EEG Clust. (2023)~\cite{ellis_convolutional_2023} & Autoencoder + $k$-means           & Cluster-specific spectral patterns            & Post-hoc & EEG brain states\\
CLAMP (2022)~\cite{bobek_enhancing_2022}               & Model-agnostic framework          & Post-hoc rules and prototypes                 & Post-hoc & Cyber-security logs\\
\bottomrule
\end{tabularx}
\end{table}

To summarize the landscape of methods,~\autoref{tab:deep_ts_xai} provides a summary of representative deep time-series clustering techniques that incorporate explainability. 
Each method is characterized by its model type, the explainability mechanism used, and example application domains or datasets.

\section{Application Domains}

Explainable deep clustering methods have been applied in various domains, where the combination of unsupervised discovery and interpretability is especially valuable, such as healthcare, finance, sensor networks, and others.
In this section, we focus on the application domains driving XAI development in time series clustering.

\subsection{Healthcare}

This is a major area driving XAI for time series. 
Patient data (vital signs, lab results, EEG signals, etc.) are rich in temporal patterns that clinicians want to cluster into meaningful states or phenotypes~\cite{manduchi_t_2021}. 
Importantly, these clusters must be interpretable to be clinically and medically useful. 
For example, ICU patient state clustering has been tackled by T-DPSOM~\cite{manduchi_t_2021} – the model clustered multivariate ICU time series into discrete health states and provided a 2D map visualization of these states. 
Clinicians could follow a patient’s trajectory on the map (e.g., moving from a stable state to a critical state), and the map’s topology gave insights into similarity of states~\cite{manduchi_t_2021}. 
Another example is resting-state EEG clustering in neurological disorders. 
Ellis et al.~\cite{ellis_convolutional_2023} clustered EEG recordings from schizophrenia patients into 8 distinct “brain states” using a deep autoencoder with k-means, and found each cluster corresponded to a different level of delta-band activity~\cite{ellis_convolutional_2023}.
By correlating cluster membership with clinical symptoms, and explaining clusters via EEG spectral features (delta power), the study provided an interpretable link between EEG patterns and disease state~\cite{ellis_convolutional_2023}. 
More generally in healthcare, explainable clustering helps in phenotype discovery (uncovering subtypes of diseases from time series biosignals) and treatment monitoring, where trust and insight are crucial. 
Methods like attention-based clustering can highlight which vital sign trends are associated with a given patient cluster (e.g., “Cluster A: patients with sustained high heart rate variability”), and prototype-based methods can present a representative patient trajectory for each cluster. 
This fosters clinician acceptance as they can verify that clusters make medical sense.

\subsection{Finance}

Financial time series (stock prices, economic indicators, transaction sequences) are often clustered to identify market regimes or customer behaviors. 
Here, explainability is needed for regulatory and strategic reasons – analysts must understand why clusters form (e.g., a group of stocks moving together due to an underlying sector trend). 
While fewer works explicitly focus on deep clustering in finance, the techniques are applicable. 
For instance, a deep clustering of multivariate stock time series might reveal clusters corresponding to “growth stocks” vs “value stocks,” and an attention mechanism could highlight that one cluster’s series are driven by certain time periods (e.g., all surged during a specific economic event). 
In customer analytics (like credit card usage patterns or mobile banking activity over time), clustering can find user segments. 
Explainable clustering would allow a bank to describe each segment (cluster) in terms of interpretable features – e.g., Cluster X: “high seasonal spending around holidays, then dormancy”, whereas Cluster Y: “consistent weekly expenditures with gradual growth.” 
By employing post-hoc methods (like decision trees on summary features such as transaction frequency, volatility, etc.), the bank can communicate these insights clearly. 
Some recent studies in financial profiling use deep autoencoders for clustering and then apply explainability methods to link clusters to demographic or behavioral features~\cite{higashi_decision_2023}, enabling domain experts to validate and act on the clusters.

\subsection{IoT and Sensor Networks}

In Internet of Things applications, large networks of sensors generate time series data (e.g., smart home energy usage, industrial machine sensor readings, traffic flows). 
Clustering such data helps identify usage patterns, anomalies, or system states. 
Explainable deep clustering plays a role in smart energy management, for example, grouping households by electricity usage patterns. 
An interpretable clustering might find a cluster of homes with “night-peaking” usage vs “daytime-peaking” usage, and XAI methods could highlight which time-of-day features distinguish these clusters. 
A recent study clustered residential electricity demand time series and emphasized interpretability by using feature selection and rule-based descriptions for each cluster~\cite{kallel_clustering_2025}. 
In industrial IoT, deep clustering might categorize machine states from multivariate sensor logs; explaining these clusters could involve showing prototypical sensor readings or pinpointing which sensor signals (vibration, temperature, etc.) are most indicative of each operational mode. 
Virtual inspection techniques (like converting sensor time series to frequency spectra) can help domain engineers understand what a cluster represents (e.g., a certain vibration frequency pattern indicating a machine fault)~\cite{vielhaben_explainable_2024}. 
Overall, in IoT scenarios, explainability ensures that the patterns discovered by unsupervised models translate into actionable insight (for engineers, operators, or consumers).

\subsection{Other domains}

Transportation and mobility data have seen deep clustering for pattern mining (e.g., grouping city traffic flow time series)~\cite{barredoarrieta_what_2019}. 
Here, explaining clusters could involve mapping them to known traffic conditions or events (rush-hour vs off-hour patterns) and using attention to show which intervals (holidays, weekends) influenced the clustering. 
Environmental science uses time series clustering for things like climate patterns or animal migration sequences; experts demand explanations like “Cluster A corresponds to El Niño years with these temperature fluctuations”~\cite{vaittinadaayar_regime_2023}.
Deep clustering can be combined with feature attribution to tie clusters to physical phenomena. 
In summary, any field dealing with complex temporal data can benefit from explainable clustering: the deep models provide power to handle nonlinear, high-dimensional sequences, while the explainability component ensures the results are trustworthy to domain experts.

\section{Research Opportunities}

The convergence of deep learning and explainable clustering for time series is a relatively new but rapidly evolving research area. 
\autoref{tab:deep_ts_xai} contrasted several methods, and here we distill a few comparative observations:

\subsection{Model Architectures and Explainability}

There is often a trade-off between model complexity and interpretability~\cite{guidotti_survey_2018}. 
Simpler architectures (or those augmented with interpretable components like SOMs~\cite{fortuin_som_2019} or shapelets) provide more straightforward explanations but might sacrifice some clustering accuracy or require more domain input. 
For instance, prototype-/shapelet-based methods offer very clear explanations (actual data patterns) yet may not capture subtle variations as well as a deep latent feature model. 
Conversely, deep models like transformers or VAEs can capture rich structure but need additional tools (attention weights, post-hoc analysis) to pry open their reasoning. 
The community is actively exploring hybrid solutions that achieve both high performance and inherent interpretability – e.g., semi-supervised approaches that use a few labeled examples or constraints to guide clusters towards human-meaningful categories~\cite{elamouri_constrained_2023}. 
These can improve both clustering relevance and ease of explanation.

\subsection{Quantifying and Evaluating Explanations}

Unlike supervised tasks, evaluating explanations for clustering is tricky due to the lack of ground truth. 
In classification, one can measure if an explanation helps predict the true label; but for clustering, “ground truth” clusters may not exist. 
Researchers have used proxy metrics – e.g., how well a post-hoc explainable model (like a decision tree) can mimic the deep model’s cluster assignments (fidelity)~\cite{schlegel_towards_2019}, or user studies to evaluate if experts agree that the explanations are meaningful~\cite{jeyakumar_how_2020}. 
An important challenge is developing objective metrics for cluster explanation quality~\cite{schlegel_introducing_2023}. 
Some works consider the compactness of explanations (e.g., few features or shapelets) and their completeness (ability to approximate the clustering)~\cite{schlegel_empirical_2020}. 
The trustworthiness of explanations is also crucial: attention scores or feature attributions need to truly reflect the model’s logic (the issue of “attention is not explanation” is debated, and methods to ensure faithfulness are needed)~\cite{schlegel_towards_2019,klein_navigating_2024}. 
Future research is exploring more effective methods to validate that an explanation for a cluster is not merely an artifact but genuinely relates to why the model grouped those series.

\subsection{Scalability and Efficiency}

Time series datasets can be large (in the number of dimensions, in the number of series, and in the length of each series). 
Deep clustering methods themselves can be computationally intensive; adding explainability can extend this. 
For example, searching for shapelets or training a global surrogate model might not scale easily to millions of time series. 
Recent works address this by dimensionality reduction (e.g., Time2Feat selecting a small set of features~\cite{bonifati_time2feat_2022}) or by focusing explanations on a subset of important clusters. 
An open challenge is to create explainable clustering pipelines that remain efficient on big data, possibly by leveraging online learning or distributed computing. 
Additionally, dynamic time series (streams that evolve) raise questions: clusters may drift over time, so explanations might need to update continuously, incorporating data stream XAI approaches~\cite{fumagalli_incremental_2023}. 
How to maintain an interpretable model in non-stationary settings is largely unexplored and needs more research.

\subsection{Generality and Domain-Specificity}

We observe that some explainable clustering methods are domain-focused (e.g., specific techniques for ECG, or special layers for speech data~\cite{bonifati_time2feat_2022}), whereas others are generic~\cite{fortuin_som_2019}. 
Domain-specific approaches can incorporate expert knowledge (like known meaningful features or patterns) to enhance interpretability – for example, focusing on frequency-domain features for audio signals because they are easier to explain to acousticians~\cite{vielhaben_explainable_2024}. 
However, too much specialization limits applicability~\cite{schlegel_visual_2023}. 
A continuing challenge is to design frameworks that are general enough to apply to many time series types, yet can ingest domain knowledge when available. 
This might entail modular designs where a model can plug in a “feature module” for interpretability (such as a Fourier or wavelet transform layer) appropriate to the domain to enhance the interpretable decision-making of models incorporating domain knowledge of experts.

\subsection{User Interaction and Human-in-the-Loop}

Explaining clusters is ultimately about helping humans make sense of data. 
There is a growing recognition that explanation is not one-size-fits-all – it should be context-dependent~\cite{theissler_explainable_2022}. 
For instance, a doctor might want a different explanation (emphasizing clinical features) than a data scientist~\cite{spinner_explainer_2019}. 
Tools like CLAMP allow users to adjust which features to consider for explanations or to provide feedback (e.g., “these two clusters should really be one” or “this feature is important, please incorporate it”).
Incorporating human feedback into deep clustering (through constraint-learning like must-link/cannot-link constraints~\cite{elamouri_constrained_2023} or interactive (visual) steering~\cite{bae_interactive_2020}) is an open avenue. 
It not only improves clustering relevance but also yields explanations aligned with human intuition (since the human helped shape the clusters)~\cite{spinner_explainer_2019}. 
The challenge is balancing human guidance with the model’s autonomy to discover unexpected patterns.

\subsection{Transparency in Model Components}

Beyond explaining the results, another challenge is making the model itself transparent. 
For example, if a deep clustering uses an LSTM encoder, can we open that LSTM and interpret its units (are certain neurons detecting specific motifs? 
Similar to detecting LSTM neurons responsible for detecting parentheisis in code~\cite{karpathy_visualizing_2015}).
Recent XAI research on interpreting recurrent or convolutional units (e.g., by clustering neuron activation patterns or visualizing filter effects) could be applied here~\cite{theissler_explainable_2022}. 
By understanding what each part of the model learns, we could provide more stable, global explanations (e.g., “this autoencoder’s latent dimension 1 corresponds to the overall trend of the series”)~\cite{schlegel_interactive_2023}. 
Achieving this remains difficult, especially as models become more complex (with hundreds of latent features), but it is a worthwhile goal for explainable deep clustering.

\section{Conclusion}


In conclusion, explainable deep clustering for time series lies at the intersection of two major trends: 
The need for unsupervised learning to handle large-scale temporal data and the growing demand for transparent, trustworthy AI. 
Recent advances, from attention mechanisms and prototype learning to shapelet discovery and post-hoc rule extraction, have begun to make clustering results more interpretable and actionable. 
Proof-of-concept applications already demonstrate value in domains like healthcare, where clinicians can inspect cluster-driving signals, and IoT, where engineers can trace anomalous device behavior.
However, significant challenges remain. 
Future work should pursue six key directions: 
(1) combining powerful deep architectures with built-in interpretability, 
(2) developing faithfulness-focused evaluation metrics for unsupervised explanations, 
(3) enabling scalable and streaming-capable explanatory clustering, 
(4) creating domain-aware explanation mechanisms, 
(5) incorporating human-in-the-loop refinement of clusters and explanations, and 
(6) improving our mechanistic understanding of model internals. 
Addressing these challenges will move the field from isolated prototypes toward reliable, explainable clustering systems that can be confidently deployed in real-world decision making.



%
%
%
\bibliographystyle{splncs04}
\bibliography{tempxai}

\begin{thebibliography}{10}
\providecommand{\url}[1]{\texttt{#1}}
\providecommand{\urlprefix}{URL }
\providecommand{\doi}[1]{https://doi.org/#1}

\bibitem{alqahtani_deep_2021}
Alqahtani, A., Ali, M., Xie, X., Jones, M.W.: Deep time-series clustering: A review. Electronics  (2021)

\bibitem{bae_interactive_2020}
Bae, J., Helldin, T., Riveiro, M., Nowaczyk, S., Bouguelia, M.R., Falkman, G.: Interactive clustering: A comprehensive review. ACM Computing Surveys  (2020)

\bibitem{barredoarrieta_what_2019}
Barredo-Arrieta, A., La{\~n}a, I., Del~Ser, J.: What lies beneath: A note on the explainability of black-box machine learning models for road traffic forecasting. IEEE Intelligent Transportation Systems Conference  (2019)

\bibitem{bibal_is_2022}
Bibal, A., Cardon, R., Alfter, D., Wilkens, R., Wang, X., Fran{\c{c}}ois, T., Watrin, P.: Is attention explanation? an introduction to the debate. Annual Meeting of the Association for Computational Linguistics  (2022)

\bibitem{bobek_enhancing_2022}
Bobek, S., Kuk, M., Szel{\k{a}}{\.z}ek, M., Nalepa, G.J.: {Enhancing cluster analysis with explainable AI and multidimensional cluster prototypes}. IEEE Access  (2022)

\bibitem{bonifati_time2feat_2022}
Bonifati, A., Del~Buono, F., Guerra, F., Tiano, D.: {Time2Feat: Learning interpretable representations for multivariate time series clustering}. VLDB Endowment  (2022)

\bibitem{dilokthanakul_deep_2016}
Dilokthanakul, N., Mediano, P.A.M., Garnelo, M., Lee, M.C.H., Salimbeni, H., Arulkumaran, K., Shanahan, M.: Deep unsupervised clustering with gaussian mixture variational autoencoders. arXiv preprint arXiv:1611.02648  (2016)

\bibitem{elamouri_constrained_2023}
El~Amouri, H., Lampert, T., Gan{\c{c}}arski, P., Mallet, C.: {Constrained DTW preserving shapelets for explainable time-series clustering}. Pattern Recognition  (2023)

\bibitem{ellis_convolutional_2023}
Ellis, C.A., Miller, R.L., Calhoun, V.D.: A convolutional autoencoder-based explainable clustering approach for resting-state eeg analysis. IEEE Engineering in Medicine and Biology Society  (2023)

\bibitem{fortuin_som_2019}
Fortuin, V., H{\"u}ser, M., Locatello, F., Strathmann, H., R{\"a}tsch, G.: {SOM-VAE: Interpretable Discrete Representation Learning on Time Series}. International Conference on Learning Representations  (2019)

\bibitem{fumagalli_incremental_2023}
Fumagalli, F., Muschalik, M., H{\"u}llermeier, E., Hammer, B.: {Incremental permutation feature importance (iPFI): towards online explanations on data streams}. Machine Learning  (2023)

\bibitem{guidotti_survey_2018}
Guidotti, R., Monreale, A., Ruggieri, S., Turini, F., Giannotti, F., Pedreschi, D.: A survey of methods for explaining black box models. ACM Computing Surveys  (2018)

\bibitem{higashi_decision_2023}
Higashi, M., Sung, M., Yamane, D., Inamuro, K., Nagai, S., Kobayashi, K., Nakata, K.: {Decision Tree Clustering for Time Series Data: An Approach for Enhanced Interpretability and Efficiency}. Pacific Rim International Conference on Artificial Intelligence  (2023)

\bibitem{hu_interpretable_2024}
Hu, L., Jiang, M., Dong, J., Liu, X., He, Z.: {Interpretable Clustering: A Survey}. arXiv preprint arXiv:2409.00743  (2024)

\bibitem{ienco_deep_2020}
Ienco, D., Interdonato, R.: Deep multivariate time series embedding clustering via attentive-gated autoencoder. Pacific-Asia Conference on Knowledge Discovery and Data Mining  (2020)

\bibitem{jeyakumar_how_2020}
Jeyakumar, J.V., Noor, J., Cheng, Y.H., Garcia, L., Srivastava, M.: {How Can I Explain This to You? An Empirical Study of Deep Neural Network Explanation Methods}. Advances in Neural Information Processing Systems  (2020)

\bibitem{kallel_clustering_2025}
Kallel, S., Amayri, M., Bouguila, N.: {Clustering and Interpretability of Residential Electricity Demand Profiles}. Sensors  (2025)

\bibitem{karpathy_visualizing_2015}
Karpathy, A., Johnson, J., Fei-Fei, L.: Visualizing and understanding recurrent networks. arXiv preprint arXiv:1506.02078  (2015)

\bibitem{klein_navigating_2024}
Klein, L., L{\"u}th, C., Schlegel, U., Bungert, T., El-Assady, M., J{\"a}ger, P.: Navigating the maze of explainable ai: A systematic approach to evaluating methods and metrics. Advances in Neural Information Processing Systems  (2024)

\bibitem{lundberg_unified_2017}
Lundberg, S., Lee, S.I.: A {{Unified Approach}} to {{Interpreting Model Predictions}}. Advances in {{Neural Information Processing Systems}}  (2017)

\bibitem{maaten_visualizing_2008}
van~der Maaten, L., Hinton, G.: {Visualizing Data Using T-{{SNE}}}. Journal of Machine Learning Research  (2008)

\bibitem{manduchi_t_2021}
Manduchi, L., H{\"u}ser, M., Faltys, M., Vogt, J., R{\"a}tsch, G., Fortuin, V.: T-dpsom: An interpretable clustering method for unsupervised learning of patient health states. Conference on Health, Inference, and Learning  (2021)

\bibitem{manduchi_dpsom_2019}
Manduchi, L., H{\"u}ser, M., Vogt, J., R{\"a}tsch, G., Fortuin, V.: {DPSOM: Deep probabilistic clustering with self-organizing maps}. arXiv preprint arXiv:1910.01590  (2019)

\bibitem{mcinnes_umap_2018}
McInnes, L., Healy, J., Melville, J.: {{Umap: {{Uniform}} Manifold Approximation and Projection for Dimension Reduction}}. arXiv preprint arXiv:1802.03426  (2018)

\bibitem{nguyen_clusformer_2021}
Nguyen, X.B., Bui, D.T., Duong, C.N., Bui, T.D., Luu, K.: Clusformer: A transformer based clustering approach to unsupervised large-scale face and visual landmark recognition. IEEE/CVF Conference on Computer Vision and Pattern Recognition  (2021)

\bibitem{paparrizos_bridging_2024}
Paparrizos, J., Yang, F., Li, H.: Bridging the gap: A decade review of time-series clustering methods. arXiv preprint arXiv:2412.20582  (2024)

\bibitem{ren_deep_2024}
Ren, Y., Pu, J., Yang, Z., Xu, J., Li, G., Pu, X., Yu, P.S., He, L.: Deep clustering: A comprehensive survey. IEEE Transactions on Neural Networks and Learning Systems  (2024)

\bibitem{ribeiro_why_2016}
Ribeiro, M.T., Singh, S., Guestrin, C.: {"Why Should {I} Trust You?": Explaining the Predictions of Any Classifier}. SIGKDD International Conference on Knowledge Discovery and Data Mining  (2016)

\bibitem{schlegel_towards_2019}
Schlegel, U., Arnout, H., El-Assady, M., Oelke, D., Keim, D.A.: {Towards a rigorous evaluation of XAI methods on time series}. IEEE/CVF International Conference on Computer Vision Workshop  (2019)

\bibitem{schlegel_time_2021}
Schlegel, U., Keim, D.A.: {Time Series Model Attribution Visualizations as Explanations}. IEEE VIS: Workshop on TRust and EXpertise in Visual Analytics (TREX)  (2021)

\bibitem{schlegel_interactive_2023}
Schlegel, U., Keim, D.A.: {Interactive Dense Pixel Visualizations For Time Series And Model Attribution Explanations}. Workshop for Machine Learning Methods in Visualisation for Big Data (MLVis)  (2023)

\bibitem{schlegel_introducing_2023}
Schlegel, U., Keim, D.A.: {Introducing the Attribution Stability Indicator: a Measure for Time Series XAI Attributions}. ECML-PKDD Workshop XAI-TS: Explainable AI for Time Series: Advances and Applications  (2023)

\bibitem{schlegel_empirical_2020}
Schlegel, U., Oelke, D., Keim, D.A., El-Assady, M.: {An Empirical Study of Explainable {{AI}} Techniques on Deep Learning Models for Time Series Tasks}. Pre-registration workshop NeurIPS  (2020)

\bibitem{schlegel_visual_2023}
Schlegel, U., Oelke, D., Keim, D.A., El-Assady, M.: {Visual Explanations with Attributions and Counterfactuals on Time Series Classification}. arXiv preprint arXiv:2307.08494  (2023)

\bibitem{spinner_explainer_2019}
Spinner, T., Schlegel, U., Sch{\"a}fer, H., {El-Assady}, M.: {{explAIner}}: {{A Visual Analytics Framework}} for {{Interactive}} and {{Explainable Machine Learning}}. IEEE Transactions on Visualization and Computer Graphics  (2019)

\bibitem{theissler_explainable_2022}
Theissler, A., Spinnato, F., Schlegel, U., Guidotti, R.: {Explainable AI for time series classification: a review, taxonomy and research directions}. IEEE Access  (2022)

\bibitem{vaittinadaayar_regime_2023}
Vaittinada~Ayar, P., Battisti, D.S., Li, C., King, M., Vrac, M., Tjiputra, J.: A regime view of enso flavors through clustering in cmip6 models. Earth's Future  (2023)

\bibitem{vielhaben_explainable_2024}
Vielhaben, J., Lapuschkin, S., Montavon, G., Samek, W.: {Explainable AI for time series via virtual inspection layers}. Pattern Recognition  (2024)

\bibitem{xie_unsupervised_2016}
Xie, J., Girshick, R., Farhadi, A.: Unsupervised deep embedding for clustering analysis. International Conference on Machine Learning  (2016)

\bibitem{ye_time_2009}
Ye, L., Keogh, E.: Time series shapelets: a new primitive for data mining. ACM SIGKDD international conference on Knowledge discovery and data mining  (2009)

\bibitem{zhou_comprehensive_2024}
Zhou, S., Xu, H., Zheng, Z., Chen, J., Li, Z., Bu, J., Wu, J., Wang, X., Zhu, W., Ester, M.: A comprehensive survey on deep clustering: Taxonomy, challenges, and future directions. ACM Computing Surveys  (2024)

\end{thebibliography}
%

\end{document}